\documentclass{article}

\usepackage{amssymb}
\usepackage{amsthm}
\usepackage{amsmath}
\usepackage{mathtools}
\usepackage{microtype}
\usepackage{graphicx}
\usepackage{subfig}
\usepackage{booktabs} 
\usepackage{sidecap}
\usepackage{algorithm}
\usepackage{algorithmic}

\newcommand{\bWW}{\mathbf{W}}

\newcommand{\bb}{\mathbf{b}}

\newcommand{\bx}{\mathbf{x}}
\newcommand{\by}{\mathbf{y}}

\newcommand{\ba}{\mathbf{a}}
\newcommand{\bv}{\mathbf{v}}
\newcommand{\bc}{\mathbf{c}}

\newcommand{\bu}{\mathbf{u}}

\newcommand{\bA}{\mathbf{A}}

\newcommand{\bC}{\mathbf{C}}

\newcommand{\bP}{\mathbf{P}}

\newcommand{\bd}{\mathbf{d}}
\newcommand{\bz}{\mathbf{z}}

\newcommand{\bq}{\mathbf{q}}

\newcommand{\hby}{\hat{\by}}

\def\<{\langle}
\def\>{\rangle}

\usepackage{hyperref}

\newtheorem{theorem}{\textbf{Theorem}}

\newtheorem{corollary}{\textbf{Corollary}}

\newtheorem{definition}{\textbf{Definition}}

\newcommand{\diag}{{\rm{diag}}}

\usepackage[accepted]{icml2019}

\icmltitlerunning{Robustness Certificates against Adversarial Examples for ReLU Networks}

\begin{document}
	
	\twocolumn[
	\icmltitle{Robustness Certificates Against Adversarial Examples for ReLU Networks}
	
	
	
	\icmlsetsymbol{equal}{*}
	
	\begin{icmlauthorlist}
		\icmlauthor{Sahil Singla}{umd}
		\icmlauthor{Soheil Feizi}{umd}
	\end{icmlauthorlist}
	
	\icmlaffiliation{umd}{University of Maryland, College Park}
	
	\icmlcorrespondingauthor{Sahil Singla}{ssingla@cs.umd.edu}
	\icmlcorrespondingauthor{Soheil Feizi}{sfeizi@cs.umd.edu}
	
	\icmlkeywords{Adversarial Examples, Robustness Certificates, Deep Neural Networks}
	
	\vskip 0.3in
	]
	
	
	
	\printAffiliationsAndNotice{\icmlEqualContribution} 
	
	\begin{abstract}
		While neural networks have achieved high performance in different learning tasks, their accuracy drops significantly in the presence of small adversarial perturbations to inputs. Defenses based on regularization and adversarial training are often followed by new attacks to defeat them. In this paper, we propose attack-agnostic robustness certificates for a multi-label classification problem using a deep ReLU network. Although computing the exact distance of a given input sample to the classification decision boundary requires solving a non-convex optimization, we characterize two lower bounds for such distances, namely the simplex certificate and the decision boundary certificate. These robustness certificates leverage the piece-wise linear structure of ReLU networks and use the fact that in a polyhedron around a given sample, the prediction function is linear. In particular, the proposed simplex certificate has a closed-form, is differentiable and is an order of magnitude faster to compute than the existing methods even for deep networks. In addition to theoretical bounds, we provide numerical results for our certificates over MNIST and compare them with some existing upper bounds. 
	\end{abstract}
	
	\section{Introduction} \label{intro}
Although neural network models have achieved state-of-the-art results on several learning tasks, in the last couple of years, researchers have demonstrated their lack of robustness with respect to adversarial perturbations. For example, in image classification, adversarial examples have been crafted to mislead the classifier while being visually indistinguishable from {\it normal} examples \cite{2014arXiv1412.6572G, 2013arXiv1312.6199S, shafahi2018adversarial}. 

In the last couple of years, a pattern has been emerged that defense mechanisms against existing attacks are often followed by stronger attacks to break them. Even detecting the presence of adversarial examples in a dataset seems to be difficult \cite{2017arXiv170204267H, 2017arXiv170507263C}. Moreover, different references have shown that adversarial examples can exist in the physical world as well \cite{sharif2016accessorize, 2016arXiv160702533K, 2017arXiv170708945E}. This can be a significant issue in deploying neural networks in applications such as self-driving cars, authentication systems, malware detection etc.
	
Studying adversarial examples for neural networks has twofold purposes: (i) devising stronger attack algorithms for crafting adversarial examples that can break the existing defense mechanisms, and (ii) developing defenses and evaluating their robustness to adversarial perturbations. In theory, the evaluation of a neural network’s robustness should be {\it agnostic} to the attack methods. However, existing methods use the distortions obtained by different attacks as an empirical robustness measure of a target neural network. As highlighted by \cite{2018arXiv180407870G}, attack based methodology provides merely an {\it upper bound} on the size of perturbation needed to fool the prediction model while security guarantees require a {\it lower bound} on the size of the adversarial perturbation. 

The robustness evaluation based on such attack approaches can cause biases in the analysis. For example,
adversarial training retrains the network by adding crafted adversarial examples using some attack methods to the training set. Although a network trained using adversarial training can be robust to the attack used to craft the adversarial examples, it can be susceptible to other types of attacks \cite{athalye2018obfuscated, athalye2017synthesizing, 2016arXiv160804644C}.

\begin{figure*}[t]
		\subfloat[h][Closest adversarial example lies inside the linear region]{\includegraphics[width=0.5\linewidth, trim={0cm 1cm 0.5cm 0cm}]{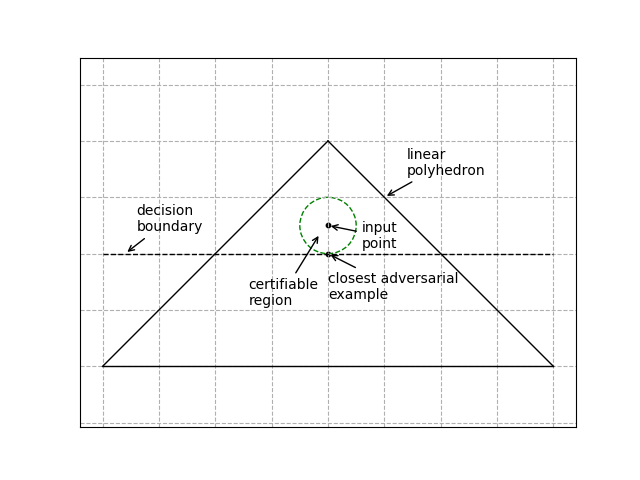}\label{fig:poly_in}}
		\hfill
		\subfloat[h][Closest adversarial example lies outside the linear region]{\includegraphics[width=0.5\linewidth, trim={0.5cm 1cm 0cm 0cm}]{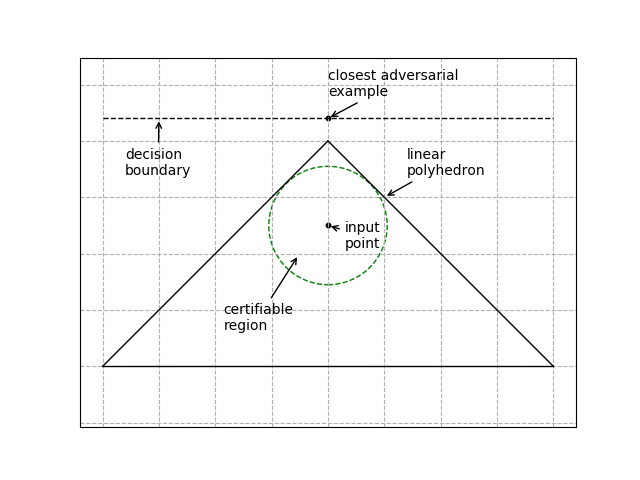}\label{fig:poly_out}}
		\caption{In our robustness certificates, we exploit the piecewise linear structure of ReLU networks, i.e. the prediction function is linear in some convex region (polyhedron) around an input point. The closest adversarial example, then, either lies inside or outside of this region. In case (a), we use the linear function to get the certificate while in case (b), we use the boundaries of the linear region to obtain a robustness certificate. In both cases, the radius of the green dotted circle gives the proposed {\it Simplex} certificate.}
		\label{fig:poly_in_out}
	\end{figure*}

In this work, we propose attack-agnostic certificates of robustness for a multi-label classification problem using a deep ReLU network. Our certificates leverage the piece-wise linear structure of deep ReLU networks and use the fact that the prediction function is linear in a polyhedron around a given sample. The key advantage of our certificate compared to other existing lower bounds (e.g. \cite{2018arXiv181100866Z, 2018arXiv180409699W}) is its extremely efficient computation even for very deep networks. Below we explain the key ideas of our proposed certificates. 

For a ReLU network, we define an {\it activation pattern} $\theta$ that represents whether or not each neuron is active (on or off) in the network. We show that for a given activation pattern $\theta$, the region in which all inputs induce that pattern on network ReLUs forms a polyhedron. We refer to this convex region by $S(\theta)$ and show that for $\bx\in S(\theta)$, the neural network is a linear function $d_\theta(.)$ where $d_\theta(\bx) = \bWW^{(\theta)}\bx + \bb^{(\theta)}$. In Section \ref{sec:relu-linear}, we explain how to efficiently compute  $S(\theta)$ and $d_\theta(.)$ for a deep ReLU network.
For a class $k$, we say that $d_{\theta}(\bx)=\bWW^{(\theta)}\bx + \bb^{(\theta)}$ defines $K-1$ decision boundaries where $K$ is the total number of classes. That is,
\begin{align*}
(\bWW^{(\theta)}_{j}-\bWW^{(\theta)}_{k})\bx + \bb^{(\theta)}_{j} - \bb^{(\theta)}_{k}=0
\end{align*}
is a decision boundary for all $ j \ne k$. $\bWW_j^{(\theta)}$ is the $j^{th}$ row of $\bWW^{(\theta)}$ and $\bb^{(\theta)}_{j}$ is the scalar in the $j^{th}$ position of $\bb^{(\theta)}$. 


For example, consider a two-layer neural network for the binary classification task where the input dimension is $D=2$ and the number of neurons in the hidden layer is $N_{1}=3$. In this case, the activation pattern $\theta$ is a binary vector of length three ($\theta \in \{0,1\}^3$.) Figure \ref{fig:poly_in_out} represents an example polyhedron defined by $\theta$ (in this case, the polyhedron is a triangle since we have three affine constraints).

Exploiting this geometric structure of the ReLU network, we characterize two lower bounds for the distance of a given input to the closest adversarial example:
	
	\begin{itemize}
		\item {\bfseries The Simplex Certificate:} This certificate applies to a ReLU network with any arbitrary depth and for a multi-label classification problem. For a given input $\bu$ (and therefore an activation pattern $\theta$ and a predicted class $k$), let $S(\theta)=\{\bx: \bP^{(\theta)} \bx + \bq^{(\theta)} \ge 0\}$ and $d_\theta(\bx) = \bWW^{(\theta)}\bx + \bb^{(\theta)}$. 

We consider two cases: The first case is when the closest adversarial example lies inside $S(\theta)$ (e.g. see Figure \ref{fig:poly_in}). In this case, 
the lower bound to the closest adversarial example is given by the minimum distance of $\bu$ to all $K-1$ decision boundaries defined by $d_{\theta}$(.):
		\begin{align*}
		&\min_{j \ne k}\frac{\mid(\bWW^{(\theta)}_{j}-\bWW^{(\theta)}_{k})\bu + (\bb^{(\theta)}_{j}-\bb^{(\theta)}_{k})\mid}{\|(\bWW^{(\theta)}_{j}-\bWW^{(\theta)}_{k})\|_{2}}
		\end{align*}
		In this case, the exact distance can be obtained by solving a linear program for every $j \ne k$. However, the above closed-form formula provides a lower bound which is very efficient to compute even for deep networks.
		
The second case is where the closest adversarial example does not lie inside $S(\theta)$ (e.g. see Figure \ref{fig:poly_out}). In this case, a lower bound to the closest adversarial example can be characterized as the distance of the point to the closest face of the polyhedron $S(\theta)$ as follows:
		\begin{align*}
		&\min_{i}  \frac{\mid\bP^{(\theta)}_{i}\bu+\bq^{(\theta)}_{i}\mid}{\|\bP^{(\theta)}_{i}\|_{2}}.
		\end{align*}
The minimum of these two quantities act as our {\it Simplex certificate}. We present details of this certificate in Section \ref{subsubsec:simplex-certificate}.

		\item {\bfseries The Decision Boundary Certificate:} Unlike the simplex certificate which can be used for a ReLU network for an arbitrary depth, this certificate applies to a two-layer ReLU network. 
			
For a given input $\bu$ with a predicted class $k$, we can write the certificate in terms of the minimum distance to all possible decision boundaries as follows:
\begin{align}\label{eq:dec-bound}
\min\limits_{j \ne k}\min\limits_{\theta^{(1)} \in \{0, 1\}^{N_{1}}}\frac{\mid (\bWW^{(\theta)}_{j}-\bWW^{(\theta)}_{k})\bu + \bb^{(\theta)}_{j}-\bb^{(\theta)}_{k} \mid}{\|\bWW^{(\theta)}_{j}-\bWW^{(\theta)}_{k}\|_{2}}
\end{align}
where ${\theta}^{(1)}$ denotes the indicator vector for the activations of the first hidden layer and $N_{1}$ is the number of hidden units in the first hidden layer.\\
We show that $\|\bWW^{(\theta)}_{j}-\bWW^{(\theta)}_{k}\|_{2} \le \|(\bWW^{(2)}_{j}-\bWW^{(2)}_{k})\|_{2}\|\bWW^{(1)}\|_{2}$
and relax $\theta^{(1)} \in \{0, 1\}^{N_{1}}$ to $\theta^{(1)} \in [0,1]^{N_{1}}$. This leads to the following lower bound:
$$\min\limits_{j \ne k}\frac{\min\limits_{\theta^{(1)}\in [0,1]^{N_{1}}}\mid (\bWW^{(\theta)}_{j}-\bWW^{(\theta)}_{k})\bu + \bb^{(\theta)}_{j}-\bb^{(\theta)}_{k} \mid}{\|(\bWW^{(2)}_{j}-\bWW^{(2)}_{k})\|_{2}\|\bWW^{(1)}\|_{2}}.$$
For a two-layer network, we show that the numerator is linear in $\theta^{(1)}$. Thus, the above optimization can be solved using convex optimization. For deeper networks, the numerator term can be a higher-order polynomial in $\theta$ making the optimization difficult to solve. Thus, for the decision boundary certificate, we only focus on two-layer neural networks. We discuss details of this certificate in Section \ref{subsubsec:decision-boundary-certificate}.
\end{itemize}
	
In what follows, we explain details of these results. All proofs have been presented in Appendix.	
	
	\section{Background and related work}\label{sec:related_work}
	
	The adversarial attacks and defenses have received significant attention from the machine learning community in the last couple of years \cite{szegedy2013intriguing,2018arXiv180205666U, goodfellow2014explaining,athalye2018obfuscated,athalye2017synthesizing,buckman2018thermometer,
		kurakin2016adversarialBIM,papernot2016distillation,zantedeschi2017efficient,papernot2016transferability,tramer2017space,carlini2016defensive, carlini2017magnet, 2018arXiv180306373K, 2017arXiv170606083M, 2017arXiv170708945E}. A wide range of defenses have been proposed to harden neural networks against adversarial attacks. However, a pattern has emerged in which the majority of adversarial defenses are broken by new attacks. For example,\cite{2016arXiv160804644C}, \cite{athalye2018obfuscated}, \cite{athalye2017synthesizing}, \cite{2018arXiv180205666U} broke several of these proposed defenses.
	
	\subsection{Creating Adversarial Examples}
	The current state-of-the-art white-box attack methods are the iterative fast gradient sign method (I-FGSM) \cite{2014arXiv1412.6572G, 2016arXiv161101236K}, DeepFool \cite{moosavi2016deepfool}, Carlini and Wagner’s attack (CW attack) \cite{2016arXiv160804644C}, elastic-net attacks to deep neural networks \cite{2017arXiv170904114C}, robust physical perturbations attack \cite{2017arXiv170708945E}, EOT attack \cite{2017arXiv170707397A}. In the white-box attacks, the network parameters are assumed to be visible to the attacker. Black-box adversarial attacks are also possible by computing universal adversarial perturbations \cite{2016arXiv161008401M}, using ensemble approaches \cite{2017arXiv171200673L}, using substitute models \cite{2016arXiv160202697P, 2018arXiv180408598I}, employing zero-order optimization-based attacks \cite{2017arXiv170803999C, 2018arXiv180707978I}.
	
	\subsection{Certifiable Defenses}
	A number of ``certifiable'' defense mechanisms have been developed for certain classifiers. \cite{raghunathan2018certified} harden a two-layer classifier using semidefinite programming, and \cite{sinha2018certifying} proposes a convex duality-based approach to adversarial training that works on sufficiently small adversarial perturbations with a quadratic adversarial loss, while \cite{kolter2017provable} considers training a robust classifier using the convex outer adversarial polytope. \cite{2018arXiv181012715G} shows how applying interval bound propagation during training, combined with MILP-based exact verification, can lead to provably robust networks. These provable defenses, although very insightful, are either restrictive or are computationally expensive. 
	
	\subsection{Theoretical Robustness Guarantees against Adversarial Examples}
	In recent years, formal verification methods were
developed to verify robustness of neural networks. Most of these methods use satisfiability modulo theory (SMT) solvers
	 \cite{2017arXiv170501320E, 2017arXiv170201135K, carlini2016towards} or Large scale Mixed integer Linear Programming (MILP) solvers \cite{2017arXiv170501040C, 2017arXiv170607351L, 2017arXiv171107356T}. However, these methods scale poorly with the number of ReLUs in a network, making them prohibitively slow in practice even for medium-sized models. \cite{2017arXiv170201135K} has illustrated the difficulty of exact verification by proving that it is NP-complete.

Some methods computes robustness certificates by computing the Lipschitz constants. For example, \cite{2013arXiv1312.6199S} evaluates the Lipschitz constant for each layer of the network and uses the product of these Lipschitz constants to demonstrate the robustness issue in neural networks. Reference  \cite{2017arXiv170508475H} derives a closed-form robustness certificate using a local Lipschitz continuous condition for a single-hidden layer feed-forward network. However, a closed-form lower bound for an arbitrary depth multi-layer-perceptron (similar to our simplex certificate) seems to be difficult to compute. \cite{2018arXiv180110578W} computes a characterization of the lower bound for distances to the closest adversarial examples. However, as highlighted by \cite{2018arXiv180407870G}, their proposed method computes an empirical estimate of this theoretical lower bound and hence is not a robustness certificate in general. \cite{2018arXiv181100866Z, 2018arXiv180409699W} compute a theoretical lower bound but it can be expensive to compute for large networks.

An {\it upper bound} to the distance of a point to its closest adversarial example can be computed using a certain attack \cite{2016arXiv160507262B}. This differs from our proposed robustness certificates because our certificate is a lower bound on the minimum distortion and is attack agnostic. Additionally, our simplex certificate is differentiable and can be computed in closed form for a given input for any deep multi-layer perceptron with ReLU activations. We show that simplex certificate is an order of magnitude faster to compute than the state of the art methods \cite{2018arXiv181100866Z, 2018arXiv180409699W}.
	
\section{Piecewise Linear Structures of ReLU Networks}\label{sec:relu-linear}
	In this section, we provide notation and definitions that will be used in characterizing our robustness certificates. To simplify the exposition, the approaches are developed under the notation of fully connected networks with ReLU activations for a multi-label classification problem.
	
\subsection{Notation}\label{subsec:lower_bound_notation}
	We consider a neural network with $M$ layers and $N_{i}$ neurons in the $i^{th}$ layer ($M\ge2$ and $i \in$ $[M]$) for a multi-class classification problem. Let $N$ be the total number of hidden neurons in the network. The number of classes is equal to $K$ (or $N_M$). The corresponding function of neural network is $f:\mathbf{R}^D \to \mathbf{R}^K$ where $D$ is the dimension of the input. 
	
	We use $\bx$ to represent an input instance in $\mathbf{R}^D$. We use $[L]$ to denote the set $\{1,\dotsc,L\}$. For an input $\bx$, we use $\bz^{(i)}(\bx) \in \mathbf{R}^{N_{i}}$ and $\ba^{(i)}(\bx) \in \mathbf{R}^{N_{i}}$ to denote the input ({\it before} applying ReLU activations) and output ({\it after} applying ReLU activations) of neurons in the $i$-th hidden layer of the network, respectively. For simplicity, we refer to $\bz^{(i)}(\bx)$ and  $\ba^{(i)}(\bx)$ as raw and activated neurons in the $i$-th layer, respectively. The raw and activated neurons in the $j$-th position of the $i$-th hidden layer are given by $\bz^{(i)}_{j}(\bx)$ and $\ba^{(i)}_{j}(\bx)$ respectively.  To simplify notation and when no confusion arises, we make the dependency of $\bz^{(i)}$ and $\ba^{(i)}$ to $\bx$ implicit. We define $\ba^{(0)}(\bx)=\bx$ and $N_{0}=D$. 
		
	With a fully connected architecture and ReLU activations, each $\bz^{(i)}$ and $\ba^{(i)}$ (for $i \in$ $[M]$) is computed using a transformation matrix $\bWW^{(i)} \in R^{N_{i} \times N_{i-1}}$ and the bias vector $\bb^{(i)} \in R^{N_{i}}$ as follows:
	\begin{align*}
	&\bz^{(i)}(\bx) = \bWW^{(i)}\ba^{(i-1)}(\bx) + \bb^{(i)}\\
	&\ba^{(i)}(\bx) = \text{ReLU}\left(\bz^{(i)}(\bx)\right) = \max\left(0, \bz^{(i)}(\bx)\right)\\
	&\ba^{(M)}(\bx) = \bz^{(M)}(\bx).
	\end{align*}
	
	The weight and bias vectors for the $j$-th row of $\bWW^{(i)}$ and $\bb^{(i)}$ are given by $\bWW^{(i)}_{j}$ and $\bb^{(i)}_{j}$, respectively. For a given input $\bx$, the vector of logits is given by,
	\begin{align*}
	f(\bx) = \bz^{(M)}(\bx).
	\end{align*}
	We use $f_{i}(\bx)$ to denote the logit for the class $i$  where $i \in [K]$. The predicted class is given by
	\begin{align*}
	\hby(\bx) = \underset{i \in [K]}{\text{argmax }} f_{i}(\bx).
	\end{align*}
	We use $\odot$ to denote the Hadamard Product. We use $\bA_{i}$ to denote the $i^{th}$ row of the matrix $\bA$. For a vector $\bv$, we use $\bv_{i}$ to denote the element in the $i^{th}$ position of the vector.  We use $\text{diag}(\bv)$ to denote the diagonal matrix formed by placing each element of $\bv$ along the diagonal.
	
	\begin{definition}[Activation pattern]\label{def:activation-pattern}
		An activation pattern $\theta$ is a set of indicator vectors for each hidden layer of the network. The indicator vector for $i^{th}$ hidden layer $(i \in [M-1])$ is denoted by $\theta^{(i)}$ and  specifies the following functional constraints: 
		
		$$\theta^{(i)}_{j}=
		\begin{cases}
		1,\quad\bz^{(i)}_{j}(\bx)\ge 0\\
		0,\quad \bz^{(i)}_{j}(\bx)< 0
		\end{cases}$$			
		We denote $\theta$ as $\theta = \{\theta^{(1)},\dotsc,\theta^{(M-1)}\}$.		
	\end{definition}
	
We say that an input $\bx$ induces an activation pattern $\theta$ in $f$ if the activation pattern defined by the neurons $\bz^{i}_{j}(\bx)$ is equal to $\theta$. Informally, an activation pattern $\theta$ represents a configuration of all ReLUs in the network as either being "on" or "off".
	
	\begin{definition}[Activation Region]\label{def:activation-region}
		For a given activation pattern $\theta$, we define the activation region $S(\theta)$ such that:
		\begin{align*}
		\bx \text{ induces }\theta \text{ in f } \iff \bx \in S(\theta)
		\end{align*}
	\end{definition}
	Thus the activation region for an activation pattern $\theta$, is the largest region such that all inputs in the region induce the activation pattern $\theta$.
	
	\begin{definition}[Decision function]\label{def:decision-function}
		For a given activation pattern $\theta$, we define the decision function $d_{\theta}:\mathbf{R}^d \to \mathbf{R}^K$ such that for every $\bx$ that induces the activation pattern $\theta$:
		\begin{align*}
		d_{\theta}(\bx)= f(\bx), \quad \text{ } \forall\text{ }\bx \in S(\theta).
		\end{align*}
	\end{definition}
	
	The decision function is the function such that for all inputs that induce a certain activation pattern, the neural network and the decision function are the same.
	
	
	\subsection{Polyhedral Structures of ReLU Networks}\label{subsec:polyhedral_structure}
	Because ReLU networks are piece-wise linear functions, they are linear in some region around a given input. In this section, we prove that for an arbitrarily deep ReLU network, the activation region for an activation pattern is a convex polyhedron and in this region, the neural network is linear. Furthermore, we derive the exact polyhedron in which the activation pattern is constant. 
	
We first explain our results for a two-layer neural network and then present our them for a neural network with an arbitrary depth. For a two-layer ReLU networks, the activation pattern $\theta$ is merely composed of one vector $\theta^{(1)}$ (since $M=2$ in this case). In this case, we have:
	
	\begin{theorem}\label{thm:activation-pattern-2layer}
		Given an activation pattern $\theta=\{\theta^{(1)}\}$ for a two-layer network, we have:
		
		(a) The activation region is $S(\theta)$ where
		\begin{align*}
		&S(\theta) = \cap_{i=1}^{i=N_{1}}S^{(1)}_{i}(\theta) \text{ , }\\
		&S^{1}_{i}(\theta) = 
		\begin{dcases} 
		\bx :  \bWW^{(1)}_{i}\bx + \bb^{(1)}_{i} \ge 0 & \text{if } \theta^{(1)}_{i}=1 \\
		\bx :  \bWW^{(1)}_{i}\bx + \bb^{(1)}_{i} < 0 & \text{if }\theta^{(1)}_{i}=0 
		\end{dcases}
		\end{align*}
		
		(b) The decision function is $d_{\theta}(.)$ where
		\begin{align*}
		&d_\theta(\bx) = \bWW^{(\theta)}\bx + \bb^{(\theta)} \text{, }\\
		&\bWW^{(\theta)} = \bWW^{(2)}\diag(\theta^{(1)})\bWW^{(1)} \text{ , } \\
		&\bb^{(\theta)} = \bWW^{(2)}\diag(\theta^{(1)})\bb^{(1)} + \bb^{(2)}
		\end{align*}
		
	\end{theorem}
	
	Note that in this case $S(\theta)$ is a polyhedron characterized using $N_1$ linear constraints (i.e. $N_1$ faces). At a vertex, D linearly independent constraints are tight (i.e. equal to zero). In total, $\binom{N1}{D}$ combinations are possible. But all the $\binom{N1}{D}$ combinations may not have D linearly independent equations and some of them may not have satisfy the other inequality constraints of the polyhedron. 
	
	A similar result can be stated for a ReLU network with an arbitrary depth (although the notation is a bit more complex than that of the two-layer case). 
	
	We present this result in the following theorem:
	
	\begin{theorem}\label{thm:activation-pattern-multilayer}
		Consider an $M$ layer $(M\ge3)$ neural network (denoted by $f(.)$) and an activation pattern $\theta=\{\theta^{(1)},\dotsc, \theta^{(M-1)}\}$.
		Consider the $M-1$ layer neural network (denoted by $g(.)$) constructed by removing the last weight layer and the last activation layer such that $$g(\bx) = \bz^{(M-1)}(\bx)\text{ }~~\forall\text{ }\bx\in\mathbf{R}^D$$ and the activation pattern $\phi=\{\theta^{(1)}, \dotsc, \theta^{(M-2)}\}$. Given the decision function $d_{\phi}(\bx) = \bWW^{(\phi)}\bx + \bb^{(\phi)}$ and the activation region $T(\phi)$ for g(.), 
		(a) the activation region for $f(.)$ is $S(\theta)$ where:
		\begin{align*}
		&S(\theta) = T(\phi) \cap S^{(M-1)}(\theta) \text{ where, }\\ 
		&S^{(M-1)}(\theta) = \cap_{i=1}^{i=N_{M-1}}S^{(M-1)}_{i}(\theta) \text{ and } \\
		&S^{(M-1)}_{i}(\theta) = 
		\begin{dcases} 
		\bx :  \bWW^{(\phi)}_{i}\bx + \bb^{(\phi)}_{i} \ge 0 & \text{if } \theta^{(M-1)}_{i}=1 \\
		\bx :  \bWW^{(\phi)}_{i}\bx + \bb^{(\phi)}_{i} < 0 & \text{if }\theta^{(M-1)}_{i}=0 
		\end{dcases}
		\end{align*}
		and (b) the decision function for $f(.)$ is $d_{\theta}(.)$ where:
		\begin{align*}
		&d_{\theta}(\bx) = \bWW^{(\theta)}\bx + \bb^{(\theta)} \text{  where, }\\
		&\bWW^{(\theta)} = \bWW^{(M)}\diag(\theta^{(M-1)})\bWW^{(\phi)}\text{ and }\\  &\bb^{(\theta)} = \bWW^{(M)}\diag(\theta^{(M-1)})\bb^{(\phi)} + \bb^{(M)}
		\end{align*}
	\end{theorem}
As a consequence of Theorem \ref{thm:activation-pattern-multilayer}, the decision function at a layer is a function of the weights and biases of all layers upto (and including) the current layer, the activation region can be constructed using the decision function at the previous layer and the indicator vector of the current layer. Moreover, the number of linear constraints needed to define the activation region $S(\theta)$ for a deep ReLU network grows linearly with respect to the depth of the network:
	\begin{corollary}\label{corollary:activation-region-decision-function}
		For an M layer network and activation pattern $\theta$, the activation region is a polyhedron with $\sum_{i=1}^{i=M-1}N_{i}$ inequalities and the decision function is linear in the input.
	\end{corollary}
	
	Thus, the activation region $S(\theta)$ is a polyhedron and can be represented in terms of linear inequalities. 
	\begin{align*}
	S(\theta) = \{\bx: \bP^{(\theta)}\bx + \bq^{(\theta)}\ge0\}
	\end{align*}
	where $\bP^{(\theta)}$ and $\bq^{(\theta)}$ are constant given $\theta$, $\bP^{(\theta)}$ is a matrix of dimensions $(\sum_{i=1}^{M-1} N_{i}) \times D$ and $\bq^{(\theta)}$ is a vector of dimension $(\sum_{i=1}^{M-1} N_{i})$. 
	
	We emphasize that the activation region may not be the largest linear region around the input point. Indeed the linear region could be larger than the activation region if the linear function remains the same in some adjoining region with a different activation pattern. However, the activation region is provably the largest region in which the activation pattern is the same.
	
Since the decision function is locally linear in $\bx$, at the $i^{th}$ layer (with the decision function $ d_{\theta}(\bx) = \bWW^{(\theta)}\bx + \bb^{(\theta)} = \bz^{(i)}(\bx)$), we can compute the $j^{th}$ row $\bWW^{(\theta)}_{j}$ using the gradient of $\bz^{(i)}_{j}$ with respect to the input, i.e., $\nabla_{\bx} \bz^{(i)}_{j}$, in any autograd software. However, current autograd implementations do not support the gradient of a vector with respect to the neuron's input, i.e. $\nabla_{\bx} \bz^{(i)}$. Thus, to compute $\bWW^{(\theta)}$, we need to call $\nabla_{\bx} \bz^{(i)}_{j}$ once for each row of $\bWW^{(\theta)}$ ($N_{i}$ times total) making the gradient based implementation expensive. To circumvent these issues, we propose an efficient iterative method to compute the activation region and decision function for a multi-layer-perceptron in Algorithm \ref{alg:compute_theta}. 	
	
	\begin{algorithm}[h]
		\caption{\\Computing Activation Region and Decision Function}
		\label{alg:compute_theta}
		\begin{algorithmic}
			\STATE {\bfseries Input:} \\
			Input point: $\bx$\\ Network weights: $\{\bWW^{(i)}, \bb^{(i)} \quad \forall \text{ } i \in \{M\}\}$
			\STATE {\bfseries Initialize:}\\
			$\ba^{(0)} \leftarrow \bx$, $\bC \leftarrow \bWW^{(1)}$, $\bd \leftarrow \bb^{(1)}$, $\bP \leftarrow \emptyset$, $\bq \leftarrow \emptyset$\\
			\FOR{$i=1$ {\bfseries to} $M-1$}
			\STATE $\bz^{(i)} \leftarrow \bWW^{(i)}\ba^{(i-1)} + \bb^{(i)}$\\
			\STATE $\theta^{(i)} \leftarrow \bz^{(i)} \ge 0$\\
			\STATE $\ba^{(i)} \leftarrow \bz^{(i)} \odot \theta^{(i)}$\\
			\STATE $\bP^{(i)} \leftarrow \text{diag}(2\theta^{(i)}-1) \bC$\\
			\STATE $\bq^{(i)} \leftarrow \text{diag}(2\theta^{(i)}-1) \bd$\\
			\STATE $\bP \leftarrow $ concat($\bP$, $\bP^{(i)}$, axis=0)\\
			\STATE $\bq \leftarrow $ concat($\bq$, $\bq^{(i)}$, axis=0)\\
			\STATE 	$\bC \leftarrow \bWW^{(i+1)}\text{diag}(\theta^{(i)})\bC$\\
			\STATE 	$\bd \leftarrow \bWW^{(i+1)}\text{diag}(\theta^{(i)})\bd + \bb^{(i+1)}$
			\ENDFOR\\
			\RETURN $(\bP, \bq), (\bC, \bd)$
		\end{algorithmic}
	\end{algorithm}
	Algorithm \ref{alg:compute_theta} computes the activation region $S(\theta)$ and the decision function $d_{\theta}(.)$ efficiently for the activation pattern $\theta$ induced by an input $\bx$ so that we can write them in terms of the returned matrices $(\bP, \bq)$, $(\bC, \bd)$:
	\begin{align*}
	&d_{\theta}(\bx) = \bC\bx + \bd\\
	&S(\theta) = \{\bx: \bP\bx + \bq\ge0\}
	\end{align*}
	Multiplication by $\text{diag}(2\theta^{(i)}-1)$ flips the sign of the row $\bC_{j}$ if $\theta^{(i)}_{j}$ is zero and keeps the same sign otherwise. Similarly, multiplication by $\text{diag}(\theta^{(i)})$ zeros out the row $\bC_{j}$ if $\theta^{(i)}_{j}$ is zero and keeps the same row otherwise. Thus Algorithm \ref{alg:compute_theta} can compute $S(\theta)$ and $d_{\theta}(.)$ efficiently in one forward pass.
	
	\section{Robustness Certificates for ReLU Networks}\label{sec:lower_bound}
	In this section, we present our robustness certificates (lower bounds of the distance between a point to the closest adversarial examples) for ReLU networks. The closest adversarial example to a given point $\bx$ is defined as follows:

	\begin{definition}[Closest adversarial example]\label{def:adv-example}
		For a given input $\bu$ with predicted class $k$, we define the closest adversarial example $\bu_{0}$ such that:
		(a) its assigned label is different than $k$, i.e.		
		\begin{align}\label{eq:cond-misclass}
		f_{k}(\bu_{0}) = f_{j}(\bu_{0})\ge f_{i}(\bu_{0}) \quad j \neq k, \forall\ i \in [K]
		\end{align}
		(b) its distance is minimum to $\bx$ compared to all vectors $\bv$ satisfying \ref{eq:cond-misclass}, i.e.	
		\begin{align*}
		\|\bu-\bu_{0}\|_{2} \le \|\bu-\bv\|_{2} 
		\end{align*}	
	\end{definition}
	
	
	\subsection{The Simplex Certificate}\label{subsubsec:simplex-certificate}
	In this section we derive a differentiable certificate against adversarial examples that can be computed for any arbitrarily deep multi-layer-perceptron. We have illustrated the key intuition behind this certificate which we refer to it as the {\it simplex certificate} in Figure \ref{fig:poly_in_out}. If the nearest adversarial example lies inside the polyhedron $S(\theta)$, we know the linear function characterizing the decision boundary and we can compute a lower bound on the distance to the closest adversarial example (Panel (a) in Figure \ref{fig:poly_in_out}). On the other hand, if the closest adversarial example lies outside the polyhedron $S(\theta)$, we can compute a lower bound using the minimum distance of the point $\bx$ to the boundaries of the Polyhedron $S(\theta)$ (Panel (b) in Figure \ref{fig:poly_in_out}). This procedure leads to the following theorem:
	
	\begin{theorem}\label{thm:simplex-certificate}
		Given a test input $\bu$ with predicted class k and the activation pattern $\theta$ it induces, let $d_\theta(\bx) = \bWW^{(\theta)}\bx + \bb^{(\theta)}$ be the decision function and let $S(\theta) = \{\bx: \bP^{(\theta)} \bx + \bq^{(\theta)} \ge 0\}$ be the activation region. We define:
		\begin{align*}
		P_{min} &= \min_{i}  \frac{\mid\bP^{(\theta)}_{i}\bu+\bq^{(\theta)}_{i}\mid}{\|\bP^{(\theta)}_{i}\|_{2}}\\
		d_{min} &= \min_{j \ne k}\frac{\mid(\bWW^{(\theta)}_{j}-\bWW^{(\theta)}_{k})\bu + (\bb^{(\theta)}_{j}-\bb^{(\theta)}_{k})\mid}{\|(\bWW^{(\theta)}_{j}-\bWW^{(\theta)}_{k})\|_{2}}
		\end{align*}\\
		Let $\bu_{0}$ be the closest adversarial example. We have that:
		\begin{align*}
		\|\bu - \bu_{0}\|_{2} &\ge \min(d_{min}, P_{min})
		\end{align*}
		Hence, $\min(d_{min}, P_{min})$ defines a provable lower bound to the closest adversarial example.
	\end{theorem}
	Note that $N_{1}$ rows of $\bP^{(\theta)}$ and $\bq^{(\theta)}$ are a function of the weights and biases of the first layer, $N_{2}$ rows of $\bP^{(\theta)}$ and $\bq^{(\theta)}$ are a function of the weights and biases of {\it both} the first and the second layers and so on. Since both $d_{min}$ and $P_{min}$ are differentiable and can be computed in a single forward pass using Algorithm \ref{alg:compute_theta}, we can use this certificate for training robust classifiers as well. In this work, however, we focus on just characterizing this quantity as a certificate for robustness for an arbitrarily deep pre-trained ReLU network. Training robust classifiers using a regularization based on the proposed simplex certificate can be an interesting direction for the future work.
	

	\subsection{The Decision Boundary Certificate}\label{subsubsec:decision-boundary-certificate}
	In this section, we derive another lower bound to the closest adversarial example. Unlike the simplex lower bound which can be used for a ReLU network for an arbitrary depth, this lower bound is characterized for a two-layer network. \\Let  $d_\theta(\bx) = \bWW^{(\theta)}\bx + \bb^{(\theta)}$ be the decision function for the activation pattern $\theta$. Then for a given input $\bu$ with a predicted class k, we can write the certificate in terms of the minimum distance to all possible decision boundaries as in \eqref{eq:dec-bound}. Due to the dependence on $\theta$ in denominator and the combinatorial constraint $\theta^{(1)} \in \{0, 1\}^{N_{1}}$, optimization \eqref{eq:dec-bound} is difficult to solve in general. To further simplify this optimization, we first show that:
\begin{align*}
{\|\bWW^{(\theta)}_{j}-\bWW^{(\theta)}_{k}\|_{2}} &= \|(\bWW^{(2)}_{j}-\bWW^{(2)}_{k})\diag(\theta^{(1)})\bWW^{(1)}\|_{2}\\ &\le\|(\bWW^{(2)}_{j}-\bWW^{(2)}_{k})\|_{2}\|\bWW^{(1)}\|_{2}
\end{align*}
W relax $\theta^{(1)} \in \{0, 1\}^{N_{1}}$ to $\theta^{(1)} \in [0, 1]^{N_{1}}$ and solve:
$$\min\limits_{j \ne k}\frac{\min\limits_{\theta^{(1)} \in [0, 1]^{N_{1}}}\mid (\bWW^{(\theta)}_{j}-\bWW^{(\theta)}_{k})\bu + \bb^{(\theta)}_{j}-\bb^{(\theta)}_{k} \mid}{\|(\bWW^{(2)}_{j}-\bWW^{(2)}_{k})\|_{2}\|\bWW^{(1)}\|_{2}}$$
Finally, we prove that for a two-layer network $\bWW^{(\theta)}_{j}$ and $\bb^{(\theta)}_{j}$ are linear in $\theta$ and hence the above optimization can be solved efficiently. We summarize our result in the following theorem:


		\begin{theorem}\label{thm:decision-boundary-certificate}
		Given a test input $\bu$ for a 2 layer neural network with predicted class k, let $d_\theta(\bx) = \bWW^{(\theta)}\bx + \bb^{(\theta)}$ be the decision function for the activation pattern $\theta$ where:
		\begin{align*}
		\bWW^{(\theta)} &= \bWW^{(2)}\diag(\theta^{(1)})\bWW^{(1)}\\ \bb^{(\theta)} &= \bWW^{(2)}\diag(\theta^{(1)})\bb^{1} + \bb^{(2)}
		\end{align*}
		Let $\bu_{0}$ be the closest adversarial example. We have that:
		\begin{align*}
		&\|\bu - \bu_{0}\|_{2} \\
		&\ge \min_{j \ne k} \frac{ \min\limits_{\theta^{(1)} \in [0, 1]^{N_{1}}} \mid (\bWW^{(\theta)}_{j}-\bWW^{(\theta)}_{k})\bu + (\bb^{(\theta)}_{j}-\bb^{(\theta)}_{k}) \mid}{{\|\bWW^{(2)}_{j}-\bWW^{(2)}_{k}\|_{2}}\|\bWW^{(1)}\|_{2}}
		\end{align*}
		(a) RHS defines a provable lower bound to the closest adversarial example (b) $\forall\ i \in \{1,..,K\}, \ \bWW^{(\theta)}_{i}$, $\bb^{(\theta)}_{i}$ are linear in $\theta^{(1)}$ and RHS can be solved using convex optimization.
	\end{theorem}
	
For deeper networks, the numerator of the decision boundary certificate will be polynomial in $\theta$ making the optimization difficult to solve. Developing efficient computational approaches for this bound for deeper networks are among interesting directions for the future work. Nevertheless, in the next section and for a two-layer network, we include a comparison of this bound against the Simplex certificate. 
	
	\begin{figure*}[t]
		\subfloat{\includegraphics[width=0.5\linewidth, trim={0cm 0cm 0cm 1.25cm}, clip]{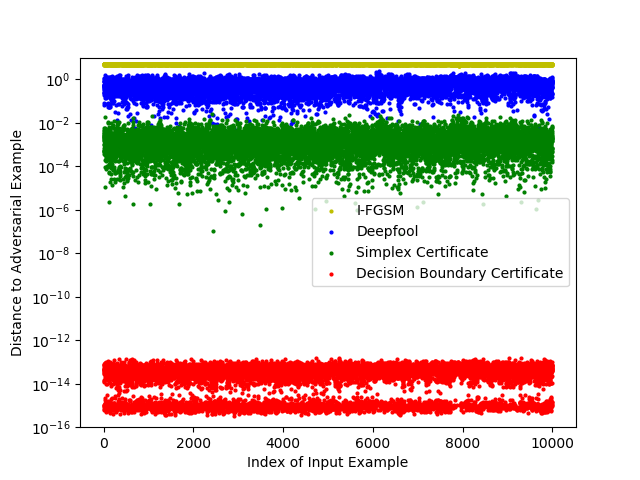}\label{fig:two-layer}}
		\hfill
		\subfloat{\includegraphics[width=0.5\linewidth, trim={0cm 0cm 0cm 1.25cm}, clip]{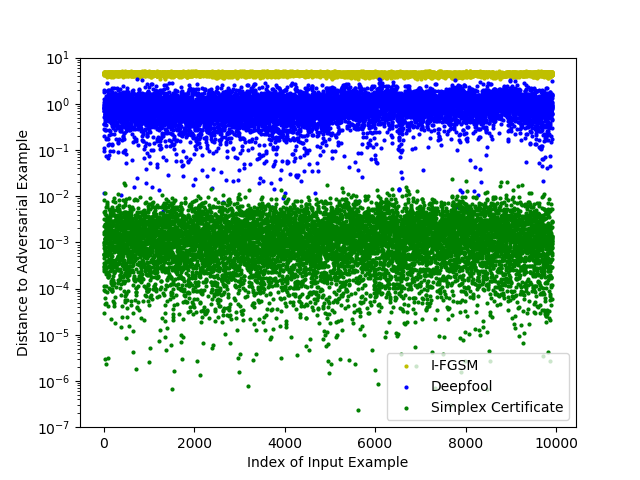}\label{fig:three-layer}}
			\caption{ (a) Comparison between simplex and decision boundary lower bounds and Iterative FGSM and DeepFool upper bounds for two-layer ReLU networks. (b) Comparison between the Simplex lower bound, and Iterative FGSM and DeepFool upper bounds for three-layer ReLU networks. Note that Decision Boundary certificate cannot be used in this case.}
	\end{figure*}	

	\begin{table*}[!h]
\caption{Running time comparison between our proposed simplex certificates and other methods. $m \times [n]$ denotes a neural network with $m$ hidden layers each with $n$ units. Running times are in seconds.}
\label{table:compare_times}
\centering
\begin{tabular}{|c|c|c|c|c|}
\hline
Config & Fast-Lin & Fast-Lip & CROWN-Ada & {\bf Simplex}\\   
\hline
\hline
	MNIST 3$\times$[1024] & 0.732 & 1.060 & 1.260 & {\bf 0.0037}\\
	MNIST 4$\times$[1024] & 1.790 & 2.580 & 3.520 & {\bf 0.0063}\\
	CIFAR 7$\times$[1024] & 12.70 & 20.90 & 20.70 & {\bf 0.0387}\\
\hline
\end{tabular}
\end{table*}

	\section{Experiments}
	In this section, we numerically assess the performance of our proposed robustness certificates. We also compare our proposed lower bounds with some existing {\it upper bounds} including the Iterative FGSM \cite{kurakin2016adversarial} and DeepFool \cite{moosavi2016deepfool}.

Note that DeepFool and Iterative FGSM provide upper bounds on the distance of a point to its closest adversarial example. Nevertheless, we compare our proposed lower bounds with these upper bounds (1) to provide a partial empirical validation of the correctness of our lower bounds, and (2) to assess the gap between the proposed lower bounds and existing upper bounds.

First, we compare the two proposed certificates, namely the simplex certificate (Section \ref{subsubsec:simplex-certificate}) and the decision boundary certificate (Section \ref{subsubsec:decision-boundary-certificate}) and DeepFool and Iterative FGSM upper bounds over a two-layer ReLU network with 1024 hidden units for an MNIST-binary classification task. The network was to classify a digit as either being $\ge$ 5 or $<$ 5.

Figure \ref{fig:two-layer} demonstrates our numerical results in this case. We observe that in all examples, the simplex certificate significantly outperforms the decision boundary one (note that a good certificate obtains large values of the lower bound). However, it may be possible that an extension of the decision boundary certificate to deeper networks performs better than the simplex certificate in some examples. We leave exploring this direction for the future work.
	
Moreover, from Figure \ref{fig:two-layer}, as expected, we observe that the values of simplex and decision boundary certificates are smaller than the values computed by Iterative FGSM and DeepFool upper bounds. We also observe that DeepFool provides a  better upper bound compared to Iterative FGSM.  

In our next experiment, we consider a deeper network than the previous case. In this case we only evaluate the performance of the simplex certificate since the decision boundary certificate is valid for two-layer networks. More specifically, we consider a three-layer MLP with ReLU activations on the MNIST dataset for digit classification. The number of hidden units were 1024 and 512 at the first and second hidden layers, respectively. 

Figure \ref{fig:three-layer} shows comparison between our simplex lower bound and DeepFool and Iterative FGSM upper bounds. We validate that our simplex certificate values are smaller than both upper bound values. Similar to the two-layer case, we observe that DeepFool provides a better upper bound than that of Iterative FGSM. In some examples, the differences between the DeepFool upper bound values and our simplex lower bound values are small, indicating the tightness of our simplex bounds in those cases.

One key advantage of the proposed simplex certificate is its efficient computation even for deep networks. This can allow using a simplex regularization in training robust classifiers. Table \ref{table:compare_times} shows a comparison of running times of our proposed simplex certificate and some other existing lower bounds including the Fast Lin, Fast-Lip proposed in \cite{2018arXiv180409699W} and CROWN-Ada proposed in \cite{2018arXiv181100866Z}. As shown in this table, our simplex certificate is an order of magnitude more efficient to compute compared to these approaches. We note that running times for other approaches reported in Table \ref{table:compare_times} are taken directly from respective references and thus can be subject to variations due to implementation differences. A comprehensive study of different aspects of these lower bounds is an interesting direction for future work.

\section{Conclusion and Future Work}
	In this paper, we characterized two robustness certificates for ReLU networks, namely the simplex certificate and the decision boundary certificate. Both of these certificates exploit the piecewise linear structure of ReLU network, i.e. for a given input, the decision boundary is linear in a polyhedron around that point. In particular, our simplex certificate is differentiable and is very efficient to compute even for deep networks.

A regularization based on the simplex certificate can be used for training neural networks to make them robust against adversarial examples. This can be an interesting direction for the future work. Another future work direction is to make the robustness lower bounds closer to the true values by exploiting piecewise linear structures of ReLU networks in neighboring polyhedra for a given point. Finally, extensions of our results to convolutional neural networks (CNNs) with ReLU activations can be another promising direction for the future work.

	

	

	
\appendix	
	
	\section{Proofs}     

\subsection{Proof of Theorem \ref{thm:activation-pattern-2layer}}\label{proof:activation-pattern-2layer}
	$(a)$ The activation pattern for $\bx$ is $\theta$ if and only if:
	\begin{align*}
	&\forall\text{ } i \in [N_{1}],\text{  }
	\begin{cases} 
	\bWW^{(1)}_{i}\bx + \bb^{(1)}_{i} \ge 0  \text{ if } \theta^{(1)}_{i}=1 \\
	\bWW^{(1)}_{i}\bx + \bb^{(1)}_{i} < 0  \text{ if }\theta^{(1)}_{i}=0 
	\end{cases}
	\end{align*}
	Thus, $S(\theta)$  gives the activation region for $\theta$.
	\begin{align*}
	&(b)\text{ }\forall\text{ }\bx \in S(\theta),\\
	&\text{ }\ba^{(1)}(\bx) =  \text{max}(\bz^{(1)}(\bx), 0)\text{ }=\text{ }\theta^{(1)} \odot \bz^{(1)}(\bx)\\
	&\text{Since }\bz^{(1)}(\bx) = \bWW^{(1)}\bx + \bb^{(1)}, \\
	&\ba^{(1)}(\bx) = \theta^{(1)} \odot (\bWW^{(1)}\bx + \bb^{(1)}) \\
	&\ba^{(1)}(\bx) = \diag(\theta^{(1)})(\bWW^{(1)}\bx + \bb^{(1)})\\
	&\ba^{(2)}(\bx) = \bz^{(2)}(\bx) = \bWW^{(2)}\ba^{(1)}(\bx) + \bb^{(2)}\\
	&\ba^{(2)}(\bx) = \bWW^{(2)}(\diag(\theta^{(1)})(\bWW^{(1)}\bx + \bb^{(1)})) + \bb^{(2)}\\
	&\ba^{(2)}(\bx) = \underbrace{\bWW^{(2)}\diag(\theta^{(1)})\bWW^{(1)}}_{\text{Weight term}}\bx \\
	&\quad\quad\quad + \underbrace{\bWW^{(2)}\diag(\theta^{(1)})\bb^{(1)} + \bb^{(2)}}_\text{Bias term}\\
	&\ba^{(2)}(\bx) = \bWW^{(\theta)}\bx + \bb^{(\theta)} \quad \text{where ,}\\
	&\bWW^{(\theta)} = \bWW^{(2)}\diag(\theta^{(1)})\bWW^{(1)} \quad \text{ and ,}\\ &\bb^{(\theta)} = \bWW^{(2)}\diag(\theta^{(1)})\bb^{(1)} + \bb^{(2)}
	\end{align*}
	Thus, $d_\theta(\bx) = \bWW^{(\theta)}\bx + \bb^{(\theta)}$ is the decision function for $\theta$.
	Since $\bWW^{(\theta)}$ and $\bb^{(\theta)}$ are constant for a given $\theta$, $\ba^{(2)}$ is linear in $\bx$ with weights $\bWW^{(\theta)}$ and bias $\bb^{(\theta)}$.

\subsection{Proof of Theorem \ref{thm:activation-pattern-multilayer}}\label{proof:activation-pattern-multilayer}
(a)\text{ We first prove } $S^{(M-1)}(\theta) \cap T(\phi) \subseteq S(\theta)$. Consider $\bx \in S^{(M-1)}(\theta) \cap T(\phi)$. Thus, $\bx \in T(\phi)$,

	\begin{align*}
	&\implies \bz^{(M-1)}(\bx) = g(\bx) = \bWW^{(\phi)}\bx + \bb^{(\phi)}\\ 
	&\ba^{M-1}(\bx) = \text{max}(\bz^{(M-1)}(\bx), 0)\\
	&\ba^{M-1}(\bx) = \text{max}(\bWW^{(\phi)}\bx + \bb^{(\phi)}, 0)
	\end{align*}
	Since $ \bx \in S^{(M-1)}(\theta)$, the indicator vector of $(M-1)^{th}$ layer is $\theta^{(M-1)}$. Thus,
	\begin{align*}
	&\implies \bx \in S(\theta)\\
	& \implies S^{(M-1)}(\theta) \cap T(\phi) \subseteq S(\theta)\\
	&\text{Now we prove }S(\theta)\ \subseteq \ S^{(M-1)}(\theta) \cap T(\phi),\\
	&\text{Consider }\bx \in S(\theta),\\
	&\text{Since }\bx \text{ induces } \theta \text{ in } f(.), 
	\text{ it }\text{ must induce } \phi \text{ in } g(.) \\
	&\text{Hence } \bx \in T(\phi),\\
	&\implies \bz^{(M-1)}(\bx) = g(\bx) = \bWW^{(\phi)}\bx + \bb^{(\phi)}\\ 
	&\text{But since } \bx \in S(\theta),\\
	&\text{The indicator vector of } (M-1)^{th} \text{ layer must be } \theta^{(M-1)}.\\
	&(M-1)^{th}\text{ indicator can be }\theta^{(M-1)} \textbf{ iff } \bx \in S^{(M-1)}(\theta)\\
	&\implies \bx \in S^{(M-1)}(\theta) \text{ and } \bx \in T(\phi)\\
	&\text{Thus, }S(\theta) \subseteq S^{(M-1)}(\theta) \cap T(\phi)\\
	&\implies S(\theta) = S^{(M-1)}(\theta) \cap T(\phi)
	\end{align*}
	\begin{align*}
	&(b)\text{ }\forall\text{ }\text{ }\bx \in S(\theta),\\
	&\ba^{(M-1)}(\bx) =\text{max}(\bz^{(M-1)}(\bx), 0)\\
	&\ba^{(M-1)}(\bx) = \theta^{(M-1)}\odot\bz^{(M-1)}(\bx)\\
	&\text{Since } \bz^{(M-1)}(\bx) = \bWW^{(\phi)}\bx + \bb^{(\phi)},\\
	&\ba^{(M-1)}(\bx) = \theta^{(M-1)} \odot (\bWW^{(\phi)}\bx + \bb^{(\phi)})\\
	&\ba^{(M-1)}(\bx) = \diag(\theta^{(M-1)})(\bWW^{(\phi)}\bx + \bb^{(\phi)})\\
	&\ba^{(M)}(\bx) = \bz^{(M)}(\bx) = \bWW^{(M)}\ba^{(M-1)}(\bx) + \bb^{(M)}\\
	&\ba^{(M)}(\bx) = \bWW^{(M)}(\diag(\theta^{(M-1)})(\bWW^{(\phi)}\bx + \bb^{(\phi)})) + \bb^{(M)}\\
	&\ba^{(M)}(\bx) = \underbrace{\bWW^{(M)}\diag(\theta^{(M-1)})\bWW^{(\phi)}}_{\text{Weight term}}\bx\\
	& \quad\quad\quad\quad+ \underbrace{\bWW^{(M)}\diag(\theta^{(M-1)})\bb^{(\phi)} + \bb^{(M)}}_\text{Bias term}\\
	& \ba^{(M)}(\bx) = \bWW^{(\theta)}\bx + \bb^{(\theta)} \quad \text{where, }\\
	& \bWW^{(\theta)} = \bWW^{(M)}\diag(\theta^{(M-1)})\bWW^{(\phi)}\text{, }\\  &\bb^{(\theta)} = \bWW^{(M)}\diag(\theta^{(M-1)})\bb^{(\phi)} + \bb^{(M)}
	\end{align*}
	
	Hence, $\bWW^{(\theta)}$ and $\bb^{(\theta)}$ are constant for a given $\theta$ and $\ba^{(M)}$ is linear in $\bx$ with weights $\bWW^{(\theta)}$ and bias $\bb^{(\theta)}$. 

\subsection{Proof of Corollary \ref{corollary:activation-region-decision-function}}\label{proof:activation-region-decision-function}
	Using Theorem \ref{thm:activation-pattern-2layer} and Theorem \ref{thm:activation-pattern-multilayer}, the $i^{th}$ hidden layer adds $N_{i}$ inequalities to the activation region and the decision function is constant given $\theta$. The proof follows using induction.

\subsection{Proof of Theorem \ref{thm:simplex-certificate}}\label{proof:simplex-certificate}
	$P_{min}$ is the minimum of the distances of the input  from all the faces of the Polyhedron $S(\theta)$. $d_{min}$ denotes the minimum of the distances of the input from all the decision boundaries defined by the decision function $d_{\theta}$.\\
	Consider two cases:\\
	Case 1: $\bu_{0}$ lies inside the polyhedron $S(\theta)$\\
	Since $\bu_{0}$ lies inside the polyhedron, 
	\begin{align*}
	f(\bu_{0})=\bWW^{(\theta)}\bu_{0} + \bb^{(\theta)}
	\end{align*}
	Since $\bu_{0}$ is the closest adversarial example, 
	\begin{align*}
	f_{j}(\bu_{0})= f_{k}(\bu_{0}) \text{ (for some }j\ne k)
	\end{align*}
	and it must lie on the decision boundary:
	\begin{align*} 
	\bWW^{(\theta)}_{j}\bu_{0} + \bb^{(\theta)}_{j} = \bWW^{(\theta)}_{k}\bu_{0} + \bb^{(\theta)}_{k} \text{ (where $j \ne k$)}
	\end{align*}
	Since $d_{min}$ is the minimum distance of $\bu$ from all such decision boundaries.
	\begin{align}\label{eq:inside_polyhedron}
	&\implies \|\bu - \bu_{0}\|_{2} \ge d_{min}
	\end{align}
	Case 2: $\bu_{0}$ lies outside the polyhedron $S(\theta)$\\
	Since $\bu_{0}$ lies outside the polyhedron, minimum distance of $\bu$ from all the faces of the polyhedron $S(\theta)$ gives a lower bound to the distance from $\bu_{0}$. Since each $\bP^{(\theta)}_{i}\bx+\bq^{(\theta)}_{i} = 0$, defines a face of the polyhedra,
	\begin{align}\label{eq:outside_polyhedron}
	\implies \|\bu - \bu_{0}\|_{2} &\ge P_{min}
	\end{align}
	Using \ref{eq:inside_polyhedron} and \ref{eq:outside_polyhedron}, 
	\begin{align*}
	\|\bu - \bu_{0}\|_{2} &\ge \min(d_{min}, P_{min})
	\end{align*}

\subsection{Proof of Theorem \ref{thm:decision-boundary-certificate}}\label{proof:decision-boundary-certificate}
	Let $\bu_{0}$ is the closest adversarial example, and $\phi$ be the activation pattern for $\bu_{0}$. Since $\bu_{0}$ must lie on a decision boundary, we assume for some l: 
	\begin{align*}
	&\bc^{(\phi)}\bu_{0} + \bd^{(\phi)}=0 \text{ where, }\\
	&\bc^{(\phi)} = \bWW^{(\phi)}_{l}-\bWW^{(\phi)}_{k}\\
	&\bd^{(\phi)} = \bb^{(\phi)}_{l}-\bb^{(\phi)}_{k}\\
	&\|\bu - \bu_{0}\|_{2} \ge \frac{\mid \bc^{(\phi)}\bu + \bd^{(\phi)} \mid}{\|\bc^{\phi}\|_{2}}
	\end{align*}
	Simplifying the denominator term $\|\bc^{(\phi)}\|_{2}$,
	\begin{align*}
	&\|\bc^{(\phi)}\|_{2} \\
	&= \|(\bWW^{(2)}_{l}-\bWW^{(2)}_{k})\diag(\phi^{(1)})\bWW^{(1)}\|_{2}\\
	&\le \|(\bWW^{(2)}_{l}-\bWW^{(2)}_{k})\|_{2}\|\diag(\phi^{(1)})\bWW^{(1)}\|_{2}\\
	&\le \|(\bWW^{(2)}_{l}-\bWW^{(2)}_{k})\|_{2}\|\diag(\phi^{(1)})\|_{2}\|\bWW^{(1)}\|_{2}\\
	&\text{Since }\|\diag(\phi^{(1)})\|_{2} = \text{max}(\phi^{(1)}) \le 1\\
	&\implies \|\bc^{(\phi)}\|_{2} \le \|(\bWW^{(2)}_{l}-\bWW^{(2)}_{k})\|_{2}\|\bWW^{(1)}\|_{2}\\
	&\frac{1}{\|\bc^{(\phi)}\|_{2}} \ge \frac{1}{\|(\bWW^{(2)}_{l}-\bWW^{(2)}_{k})\|_{2}\|\bWW^{(1)}\|_{2}}\\
	&\text{Since } \|\bu - \bu_{0}\|_{2} \ge \frac{\mid \bc^{(\phi)}\bu + \bd^{(\phi)} \mid}{\|\bc^{\phi}\|_{2}}\\
	&\text{Substituting } \frac{1}{\|\bc^{(\phi)}\|_{2}},
	\end{align*}
	\begin{align*}
	&\|\bu - \bu_{0}\|_{2} \\
	&\ge \frac{\mid \bc^{(\phi)}\bu + \bd^{(\phi)} \mid}{\|(\bWW^{(2)}_{l}-\bWW^{(2)}_{k})\|_{2}\|\bWW^{(1)}\|_{2}}\\
	&= \frac{\mid (\bWW^{(\phi)}_{l}-\bWW^{(\phi)}_{k})\bu + \bb^{(\phi)}_{l}-\bb^{(\phi)}_{k} \mid}{\|(\bWW^{(2)}_{l}-\bWW^{(2)}_{k})\|_{2}\|\bWW^{(1)}\|_{2}}\\
	&\ge \min\limits_{j \ne k}\frac{\mid (\bWW^{(\phi)}_{j}-\bWW^{(\phi)}_{k})\bu + \bb^{(\phi)}_{j}-\bb^{(\phi)}_{k} \mid}{\|(\bWW^{(2)}_{j}-\bWW^{(2)}_{k})\|_{2}\|\bWW^{(1)}\|_{2}}\\
	&\ge \min\limits_{j \ne k}\frac{\min\limits_{\theta^{(1)}_{i} \in \{0, 1\} }\mid (\bWW^{(\theta)}_{j}-\bWW^{(\theta)}_{k})\bu + \bb^{(\theta)}_{j}-\bb^{(\theta)}_{k} \mid}{\|(\bWW^{(2)}_{j}-\bWW^{(2)}_{k})\|_{2}\|\bWW^{(1)}\|_{2}}\\
	&\ge \min\limits_{j \ne k}\frac{\min\limits_{0 \le \theta^{(1)}_{i} \le 1}\mid (\bWW^{(\theta)}_{j}-\bWW^{(\theta)}_{k})\bu + \bb^{(\theta)}_{j}-\bb^{(\theta)}_{k} \mid}{\|(\bWW^{(2)}_{j}-\bWW^{(2)}_{k})\|_{2}\|\bWW^{(1)}\|_{2}}
	\end{align*}
	Since $\bWW^{(\theta)}_{j}$ and $\bWW^{(\theta)}_{k}$ are row vectors, we can simplify,
	\begin{align*}
	\bWW^{(\theta)}_{j} &= \bWW^{(2)}_{j}\diag(\theta^{(1)})\bWW^{(1)} \\
	& = (\theta^{1})^T\diag(\bWW^{(2)}_{j})\bWW^{(1)},\\
	\bb^{(\theta)}_{j} &= \bWW^{(2)}_{j}\diag(\theta^{(1)})\bb^{(1)} + \bb^{(2)}_{j}\\ 
	&= (\theta^{(1)})^T\diag(\bWW^{(2)}_{j})\bb^{(1)} + \bb^{(2)}_{j}
	\end{align*}
	Similarly for $\bWW^{(\theta)}_{k}$ and  $\bb^{(\theta)}_{k}$.\\
	Since $\bWW^{(\theta)}_{j}$, $\bb^{(\theta)}_{j}$ and $\bWW^{(\theta)}_{k}$, $\bb^{(\theta)}_{k}$ are linear in $\theta$, RHS can be solved using convex optimization.
	
\section{Details of Experiments}

\subsection{Details of Experiments reported in Figure \ref{fig:two-layer}}
Hyper-parameters used in this experiment are reported in Table \ref{tab:figure_2a}.
\begin{table}[!h]
	\caption{Hyper-parameters used in experiments of Figure \ref{fig:two-layer}}
	\centering
	\begin{tabular}{c c}
		\hline
		Parameter & Config \\   
		\hline
		\hline
		Optimizer & Adam \\
		Network architecture & [784, 1024, 2] \\
		Batch size & $64$\\
		Number of epochs & $20$\\
		Learning rate & 0.001\\
		Initialization & Glorot\\
		\hline
	\end{tabular}
	\label{tab:figure_2a}
\end{table}

\subsection{Details of Experiments reported in Figure \ref{fig:three-layer}}
Hyper-parameters used in this experiment are reported in Table \ref{tab:figure_2b}.
\begin{table}[!h]
	\caption{Hyper-parameter used in experiments of Figure \ref{fig:three-layer}}
	\centering
	\begin{tabular}{c c}
		\hline
		Parameter & Config \\   
		\hline
		\hline
		Optimizer & Adam \\
		Network architecture & [784, 1024, 512, 10] \\
		Batch size & $64$\\
		Number of epochs & $20$\\
		Learning rate & 0.001\\
		Initialization & Glorot\\
		\hline
	\end{tabular}
	\label{tab:figure_2b}
\end{table}

\subsection{Details of Experiments reported in Table \ref{table:compare_times} in the Main text}
Hyper-parameters used in this experiment are reported in Table \ref{tab:compare_times_hyp}.
\begin{table}[!h]
	\caption{Hyper-parameter used in experiments of Table \ref{table:compare_times} in the main text}
	\centering
	\begin{tabular}{c c}
		\hline
		Parameter & Config \\   
		\hline
		\hline
		Optimizer & Adam \\
		Batch size & $64$\\
		Number of epochs & $20$\\
		Learning rate & 0.001\\
		Initialization & Glorot\\
		\hline
	\end{tabular}
	\label{tab:compare_times_hyp}
\end{table}

\end{document}